\documentclass[twocolumn,10pt]{asme2e}

\usepackage{graphicx}
\usepackage{enumerate}
\usepackage{multicol}
\usepackage{xcolor}
\usepackage{amsmath}
\usepackage{url}
\confshortname{MSEC 2021}
\conffullname{the ASME 2021 16th International \\ Manufacturing Science and Engineering Conference}

\confdate{June 21-25}
\confyear{2021}
\confcity{Virtual}
\confcountry{Online}

\papernum{MSEC2021-XXXXX}

\title{DRAFT: MERGING SUBJECT MATTER EXPERTISE and DEEP CONVOLUTIONAL NEURAL NETWORK FOR STATE-BASED ONLINE MACHINE-PART INTERACTION CLASSIFICATION}

\author{Hao Wang, Yassine Qamsane, James Moyne, Kira Barton
    \affiliation{
	Department of Mechanical Engineering\\
	University of Michigan\\
	Ann Arbor, Michigan 48109\\
    Email: \{haowwang,yqamsane,moyne,bartonkl\}@umich.edu
    }	
}

\begin{document}

\maketitle    

\begin{abstract}
{\it Machine-part interaction classification is a key capability required by Cyber-Physical Systems (CPS), a pivotal enabler of Smart Manufacturing (SM). While previous relevant studies on the subject have primarily focused on time series classification, change point detection is equally important because it provides temporal information on changes in behavior of the machine. In this work, we address point detection and time series classification for machine-part interactions with a deep Convolutional Neural Network (CNN) based framework. The CNN in this framework utilizes a two-stage encoder-classifier structure for efficient feature representation and convenient deployment customization for CPS. Though data-driven, the design and optimization of the framework are Subject Matter Expertise (SME) guided. An SME defined Finite State Machine (FSM) is incorporated into the framework to prohibit intermittent misclassifications. In the case study, we implement the framework to perform machine-part interaction classification on a milling machine, and the performance is evaluated using a testing dataset and deployment simulations. The implementation achieved an average F1-Score of 0.946 across classes on the testing dataset and an average delay of 0.24 seconds on the deployment simulations.}
\end{abstract}



\section*{INTRODUCTION}

Cyber-Physical Systems (CPS), defined as the integration of physical assets with computing and communication technologies  \cite{cps_def}, are becoming omnipresent in Smart Manufacturing (SM). One notable characteristic of CPS is that the physical asset is monitored, classified, controlled, and coordinated by its cyber counterpart to achieve more efficient performance \cite{cps_monitor_control_coordinate}, and hence monitoring and classification are fundamental requirements of CPS. One important capability is machine-part interaction classification, which can be defined as the continuous characterization of the machine's effects on the work-piece in real-time \cite{gos}. Machine-part interaction classification is of special interest, because it provides information to address a number of manufacturing needs such as machine fault detection \cite{gos}, \cite{condition_monitoring_cnn} and root cause analysis \cite{context_sensitive_modeling}. The outputs of a machine-part interaction classification system are referred to as the interactive states of the machine, which are used to characterize the interactions between the machine and the work-piece.

In principle, machine-part interaction classification involves the identification of interactive states and the transitions between interactive states in real-time. The general concept of machine-part interaction classification has been used to formulate problems in other fields such as construction \cite{construction_framework}, \cite{construction_cnn_rnn}, \cite{construction_data_augmentation} and health care \cite{activity_deep_recurrent}, \cite{activity_sota}. 
While researchers have developed and tailored solutions to specific problems,  the vast majority of solutions share similar detection principles and classification algorithms. Furthermore, one can observe a common trend that deep learning techniques are becoming more widely used  \cite{condition_monitoring_cnn}, \cite{construction_data_augmentation}, \cite{activity_deep_recurrent}. Although deep learning models have achieved promising results in the manufacturing sector, they should be used with caution and proper constraints. The task of machine-part interaction classification can be conveniently formulated as a sequence of classification problems, which are well suited for deep learning models. However, classifications at different time instances are not independent of each other. As a result, a stand-alone deep learning model, regardless of its performance, is not sufficient to capture the underlying dynamics in the data. Furthermore, without proper error checking mechanisms, a deep learning model can be prone to misclassifications due to noise in the data. To construct a robust and reliable machine-part interaction classification system, we need to incorporate proper error checking mechanisms into the solution.  

After reviewing the state of the art in machine-part interaction classification and similar domains, we identify two gaps in the literature. First, a vast majority of solutions are designed to perform time series classification on steady-state behaviors, without addressing transitions between the states. While this approach is valid for a wide range of applications, it is insufficient for classifying machine-part interactions in the manufacturing sector. We argue that the information about when transitions occur is valuable for addressing manufacturing needs such as anomaly detection and root cause analysis, because temporal information of transitions can help partition useful time-series signals based on the machine-part interactions \cite{gos}. Other software and modules in the SM eco-system can leverage the partitioned signals to address corresponding manufacturing needs. Second, most existing solutions do not explicitly incorporate Subject Matter Expert (SME) knowledge. Little information about what SME knowledge is included or how to incorporate SME knowledge systematically is provided. This information is critical for adopting the solution to a specific machine or manufacturing system \cite{sme_dt}. 

We address the gaps in the literature by introducing an integrated framework featuring both deep learning techniques as well as SME knowledge for online state-based machine-part interaction classification. The first contribution of this paper is a framework that is capable of classifying the interactive states of the machine and the part and detecting transitions between these states. Our proposed deep convolutional neural network architecture explicitly exploits local connectivity of time series data in the temporal domain, while maintaining computational efficiency. The second contribution is the incorporation of SME knowledge into a data-driven approach to improve the performance of the proposed classification framework, especially with the introduction of error checking mechanisms, and how SME knowledge can be integrated systematically into the system. 

The remaining sections of the paper are organized as follows. An introduction to neural networks and a review of classification systems are presented in Section II. In Section III, an online machine-part interaction classification framework, featuring both SME and deep learning techniques, is presented. The proposed classification system is implemented for a milling machine, and the system is evaluated in terms of testing performance as well as deployment simulation performance in Section IV. Lastly, Section V provides conclusions and some future directions of this research. 

\section*{BACKGROUND}

\subsection*{Neural Networks}

A Neural Network (NN), in its most generic form, is a stack of layers that consists of computational units called neurons. Unlike traditional machine learning techniques, such as Support Vector Machine and Decision Tree, NN does not rely on human-engineered features and is capable of extracting highly complex features. Due to recent development in processing architecture like the Graphical Processing Unit, and the ever-increasing availability of data, NN-based models have achieved groundbreaking performance in domains such as computer vision and speech recognition. However, NNs are not preferred for some applications due to their poor interpretability, i.e. black-box models. 

Convolutional Neural Network (CNN) is one of the most popular variants of NN. CNN utilizes convolutional layers that perform convolutions on the input. In the context of pattern recognition of time series data, convolution is defined as follows

\begin{equation}
    s(t) = (x*\omega)(t) = \sum_a x(a)\omega(t-a)
\end{equation}

where $x(t)$ is the input, $\omega(t)$ is the kernel, and $s(t)$ is called the feature map \cite{dl_goodfellow}. For the purpose of this paper, we can understand convolution as the operation of sliding the kernel along the input and performing dot product between the kernel, a learnable filter capturing local features, and the overlapped section of the input \cite{tsc_review}. Convolutional layers are designed to capture local connectivities of the inputs, and this property makes CNN one of the most effective models for pattern recognition \cite{deep_learning_review}.

\subsection*{Classification Systems}

Classification systems are widely used in a number of domains, such as manufacturing \cite{dt_framework}, construction \cite{construction_framework}, and health care \cite{activity_sota}. While subjects of monitoring tasks differ from case to case, NN-based approaches have dominated the research landscape in recent years. This is not surprising because most monitoring tasks involve classifying segments of data, and NN-based models have succeeded in many classification tasks. 

In \cite{condition_monitoring_cnn}, Janssens et al. performed bearing condition and fault monitoring and classification using three different machine learning models: random forest classifier, support vector machine, and CNN. The models were tasked to classify the input into eight bearing conditions, and the CNN outperformed other models by at least 6\%. The CNN used in the study was composed of one convolutional layer and one fully connected layer. In another study, Slaton et al. used an NN-based model to classify activities of excavators \cite{construction_cnn_rnn}. The model used in this study is adapted from \cite{activity_sota}, and was originally designed for human activity monitoring. With some modifications, the model achieved an average F1-score of 0.78 in the classification task involving seven excavator activities. Unlike the model in \cite{condition_monitoring_cnn}, the model in this study utilizes a much deeper architecture featuring four convolutional layers and two recurrent dense layers.

NN-based approaches have achieved great success in classification, including various monitoring related tasks. However, existing solutions cannot be readily adapted to perform machine-part interaction classification, which requires simultaneous interactive state classification and transition detection.  

\section*{REQUIREMENTS OF A MACHINE-PART INTERACTION CLASSIFICATION SYSTEM}

The first requirement for a machine-part interaction classification system is being able to classify interactive states and detect transitions between interactive states. Most frameworks, as presented in the background section, are designed to classify steady-state behaviors, which exhibit consistent machine characteristics over a period of time. Such a design works well if the transitions between the states are considered irrelevant to the classification task. Unlike the applications in equipment or human activity classification, transitions between interactive states are important for classifying machine-part interactions, because the vast majority of Computer Numerical Control (CNC) machine operations are precisely predefined in their G-Code and need to be executed exactly. Without pinpointing temporal occurrences of transitions between interactive states, the monitoring system may fail to provide adequate information on how the interactive state of the machine changes over time. While G-Code provides information about actions taken by the machine, interactive states can not be readily determined from G-Code. The monitoring system needs to infer the interactive state from sensory inputs.

The second requirement is that the system must not be allowed to have a delay time greater than $\epsilon$, the maximum delay determined by SMEs. Since the proposed classification framework occurs in real-time, the delay caused by the system directly impacts how quickly control and coordination actions can take place. Furthermore, an excessive delay can derail the entire classification effort and render the classification system useless.


\section*{METHODOLOGY}

\begin{figure}[t]
    \centering
    \includegraphics[width = \linewidth]{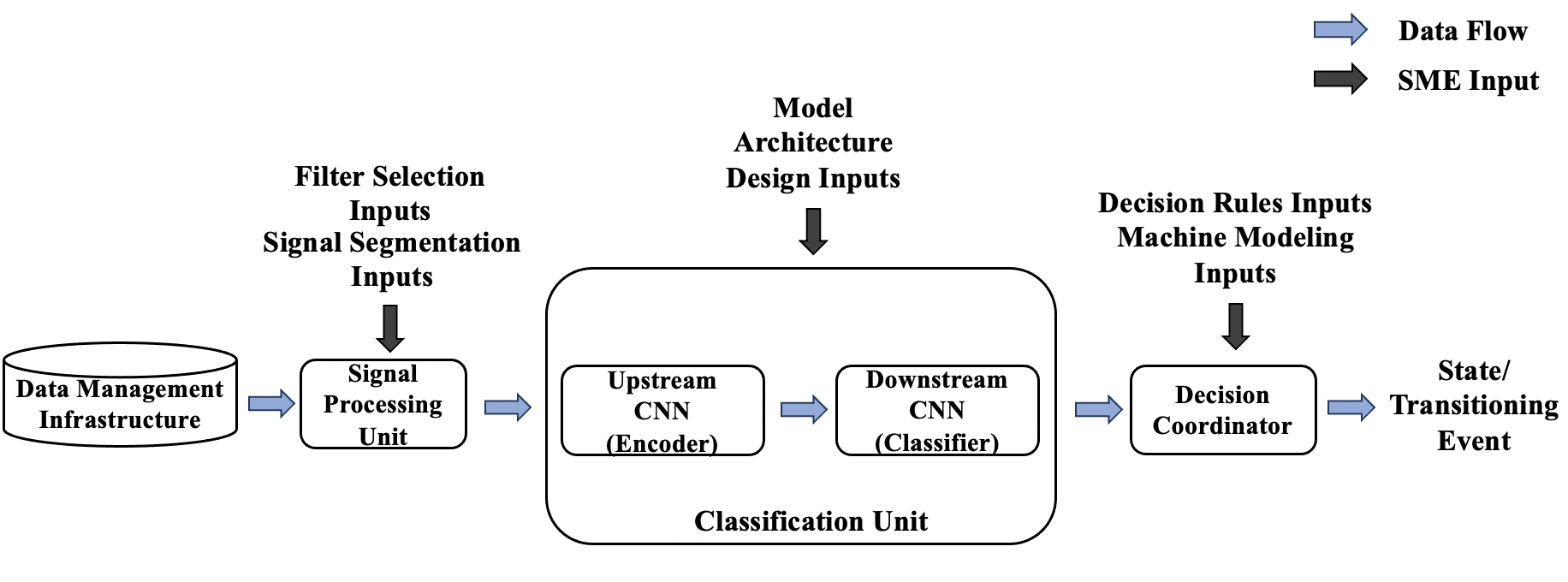}
    \caption{FRAMEWORK WORKFLOW WITH SME INVOLVEMENTS}
    \label{fig:framework_workflow}
\end{figure}

In this section, we present an online machine-part interaction classification framework. At a high level, the framework takes in a continuous stream of sensory inputs, partitions the inputs, classifies the partitioned inputs to either an interactive state or a transitioning event, and finally determines and outputs the interactive state of the machine. The machine is modeled as a Finite State Machine (FSM), whose state transitions reflect changes of the interactive state of the machine. The framework consists of three components: signal processing unit, CNN classification unit, and decision coordinator. The sliding window technique is used to partition time-series data in real-time. The components of the framework are presented in detail, followed by a brief discussion of how SME knowledge is incorporated into the framework. The framework, along with SME involvement, is illustrated in Fig. \ref{fig:framework_workflow}.  

\subsection*{Signal Processing Unit}

\begin{figure}[t]
    \centering
    \includegraphics[width = \linewidth]{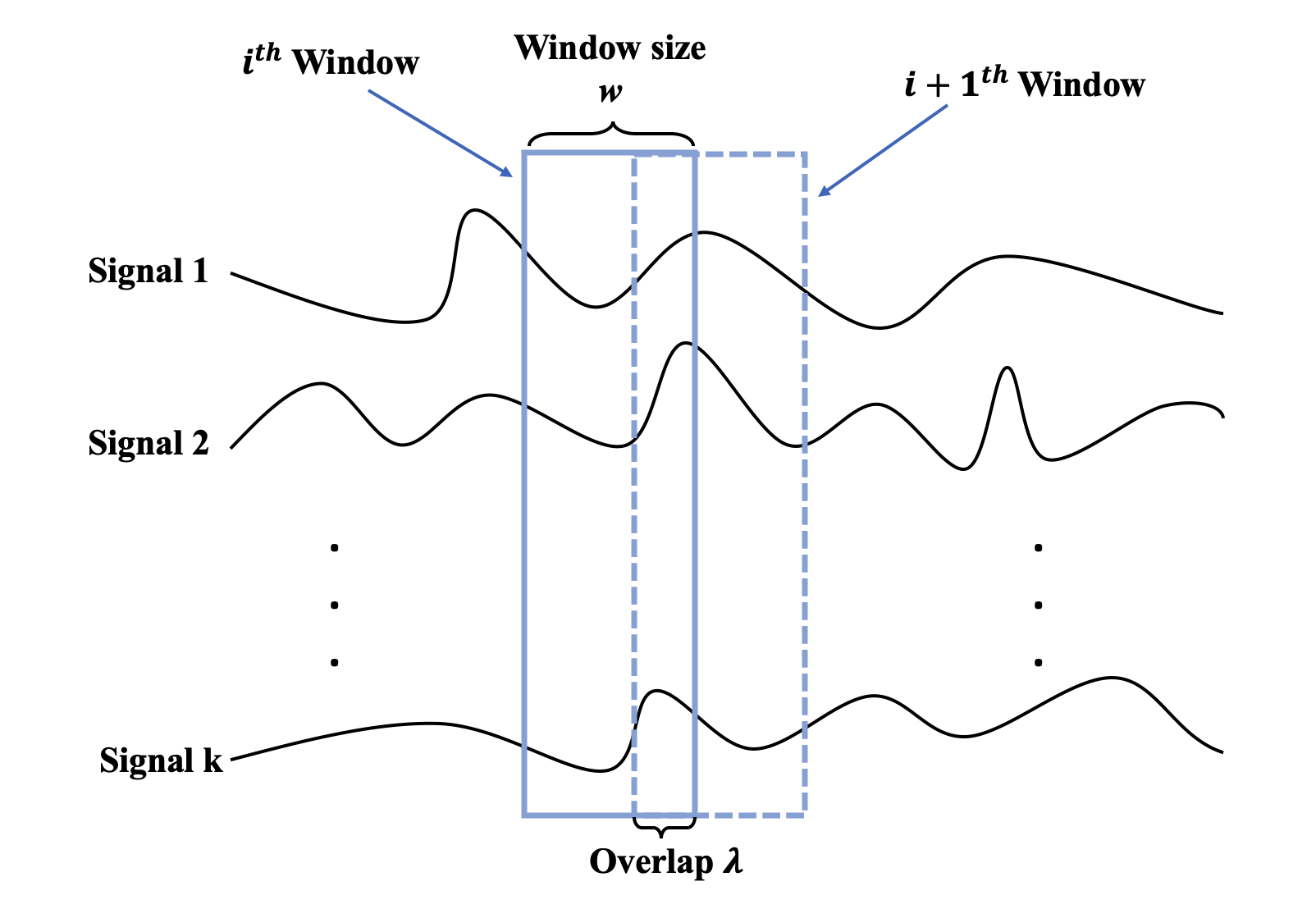}
    \caption{SLIDING WINDOW ILLUSTRATION}
    \label{fig:sw}
\end{figure}

To classify the machine-part interaction state in real-time, the continuous stream of data needs to be partitioned into segments that can be processed by the classification system. For online applications, the sliding window technique is the preferred option \cite{window_size_impact}. Two parameters, window size and overlap between windows, are determined by SMEs based on characteristics of the signals and time scale of the machine operations. The window size determines how much context is used in classification, and intuitively, larger windows can make the system more robust to noise. The overlap between windows dictates the resolution of the monitoring system; more overlap leads to better resolution (higher sensitivity for transition detection), but demands more computational resources. The sliding window technique used in this work is illustrated in Fig. \ref{fig:sw}.

Another important aspect of signal processing is denoising because it can potentially improve the performance of the classification model. There are many denoising techniques available, and the SME is in charge of selecting the appropriate techniques and relevant details, such as coefficients of filters.
 
\subsection*{Classification Unit}
\begin{figure}[t]
    \centering
    \includegraphics[width =\linewidth]{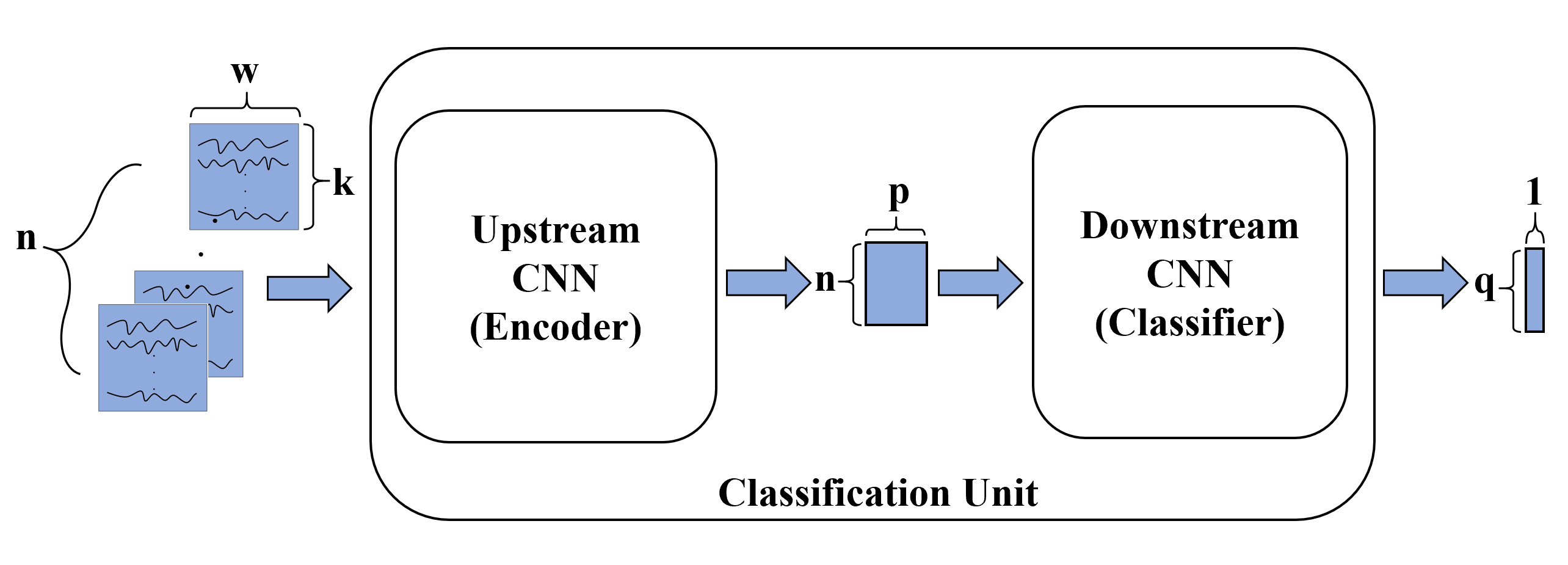}
    \caption{CLASSIFICATION UNIT ILLUSTRATION}
    \label{fig:classification_flow}
\end{figure}

The unit is a neural network composed of two CNNs, and its basic architecture is shown in Fig. \ref{fig:classification_flow}. The primary motivation for using such a design is that the class scores trajectory, obtained from a CNN trained to recognize steady-state behaviors, can well document transitions between steady states. A sequence of discrete instances of the class scores trajectory cannot only provide information for classifying steady-state behaviors, but it also indicates transitions between states. Hence, the upstream CNN is designed to encode the signal window (segment) at a time instance into class scores, and its downstream counterpart performs classification based on the sequence of class scores. While the classification process involves the entire sequence of class scores, the result only applies to the most recent window. In other words, the classification of the current window depends on how class scores change over a fixed amount of time. 

The upstream CNN acts as the encoder. The size of each window is $k \times w$, and each window is overlapped by $\lambda$ data points, as shown in Fig. \ref{fig:sw}. The input to the upstream CNN is a sequence of $n$ windows, and the input is of size $n \times k \times w$. The upstream CNN produces a $p-$class score output for each of the $n$ windows, and hence the output is of size $n \times p$, where $p=$ number of interactive states. 

The downstream CNN acts as the classifier, which transforms the $n \times p$ input into $q-$class scores, where $q = $ number of interactive states + number of transitioning events. The $q-$class score is the end output of the classification unit, and it is then fed to the decision coordinator.

Even though the classification unit does not require human-engineered features as inputs, features are still explicitly generated as an intermediate output and used for subsequent classification, as required by the encoder-classifier architecture. Other architectures such as single-stage CNNs can potentially perform as well as the purposed architecture, but the purposed architecture generates and utilizes highly compact intermediate feature representation. This architecture is easier to train and more efficient in real-time because of the compact intermediate feature representation, especially when dealing with high-dimensional data. 

\subsection*{Decision Coordinator}

By design, the framework is capable of mapping out the entire state trajectory and detecting transitions. However, without proper error checking mechanisms or restrictions imposed on state transitioning, the system would suffer from robustness issues in deployment. Two major issues have been identified during this study:
\begin{enumerate}
    \item The machine jumps from state A to state B without transitions being detected.
    \item The machine transitions from state A to state B with an impermissible transition being detected.
\end{enumerate}

To address both issues, the machine is modeled as a FSM $G = (X,E,f,\Gamma,x_o)$ \cite{discrete_event_system_book}, where $X$ is the state set, $E$ is the event set, $f$ is the transition function, $\Gamma$ is the active event function, and $x_o$ is the initial state. The definition of $G$ is determined by an SME. 

Under this modeling formalism, the output from the classification unit formally takes one of two forms: \emph{an interactive state or a transitioning event}, depending on which class has the highest class score. By definition, the transition $x_i \xrightarrow{} x_{i+1}$ cannot be triggered unless $e_i \in \Gamma(x_i)$ is present, and hence the active event function helps resolve issue 1 mentioned above by requiring a transitioning event.

Issue 2 is addressed through the transition function $f$. The formalism requires $f(x_i,e_i) = x_{i+1}$ to be defined in order for the transition $x_i \xrightarrow{} x_{i+1}$ to take place. A memory retaining device is utilized to capture $x_i,x_{i+1}$, and $e_i$, and the definition $f(x_i,e_i) = x_{i+1}$ is checked whenever a transitioning event $e_i$ is detected. This error checking mechanism is a primary motivation for requiring the classification unit to classify steady state behaviors in addition to transition events. 

At its core, the decision coordinator is a FSM implemented to perform error checking and maintain state information of the machine during operation. If either of the issues occurs in deployment, the decision coordinator will reject the proposed state transition and record the incident. The recorded incidents are added to the training dataset, and the classification unit is retrained on the new training dataset after a period of deployment.

\subsection*{Merging the Data Driven Approach with SME Knowledge}

SME knowledge, though not appropriately acknowledged, is often utilized in relevant studies. Design details, such as the architecture of the system and denoising techniques, are determined by SMEs. To facilitate adaptation and further development of the framework, we present a summary of how SME knowledge is incorporated into the framework.

While a deep learning model does not require SME-engineered features, its performance is likely affected by the quality of its input. SME knowledge is critical in selecting and packaging inputs of the monitoring system. It is a SME's responsibility to select trustworthy and distinctive signals that can reliably capture the dynamics of the machine. Furthermore, SMEs should be in charge of tuning some parameters, such as the window size $w$, sequence length $n$, and overlap $\lambda$, according to the requirements of the system and characteristics of the machine and the selected signals.

The most significant involvement of SMEs in this framework is the error checking mechanism introduced in the previous subsection. The SME-defined FSM can help remedy robustness issued it can markedly improve the performance of the classification system. Additionally, violations of the restrictions imposed by the FSM are recorded and reported to SMEs for further analysis and adjustment. The integration of SME knowledge within the framework is summarized in Fig. \ref{fig:framework_workflow}. 

\section*{IMPLEMENTATION and EVALUATION}
In this section, a machine-part interaction classification system is implemented, according to the proposed framework, to monitor a milling machine. The detailed architecture of the system is presented. The system's performance is evaluated using a testing dataset and five deployment simulations. Lastly, we provide a brief discussion on the performance of the implemented system, and how it compares to the standard method presented in the background section.

\subsection*{Problem Statement}
The case study is a simulation study and is carried out using the NASA Milling Dataset \cite{nasa_dataset}, which features 167 milling trials of 13 different sets of operation conditions (feed rate, materials, and depth of cut). The monitoring system is tasked with partitioning each trial's DC spindle current data, which is sampled at 250Hz and has a length of 9000 data points, into the following four sequential interactive states.
\begin{enumerate}[1.]
    \item \emph{No interaction}: The endmill spins freely and makes no contact with the work-piece.
    \item \emph{Entry}: The endmill starts to make contact with the work-piece.
    \item \emph{Constant milling}: The endmill makes constant contact with the work-piece.
    \item \emph{Exit}: The endmill gradually makes less contact with the work-piece.
\end{enumerate}

\begin{figure}[t]
    \centering
    \includegraphics[width = 0.45\textwidth]{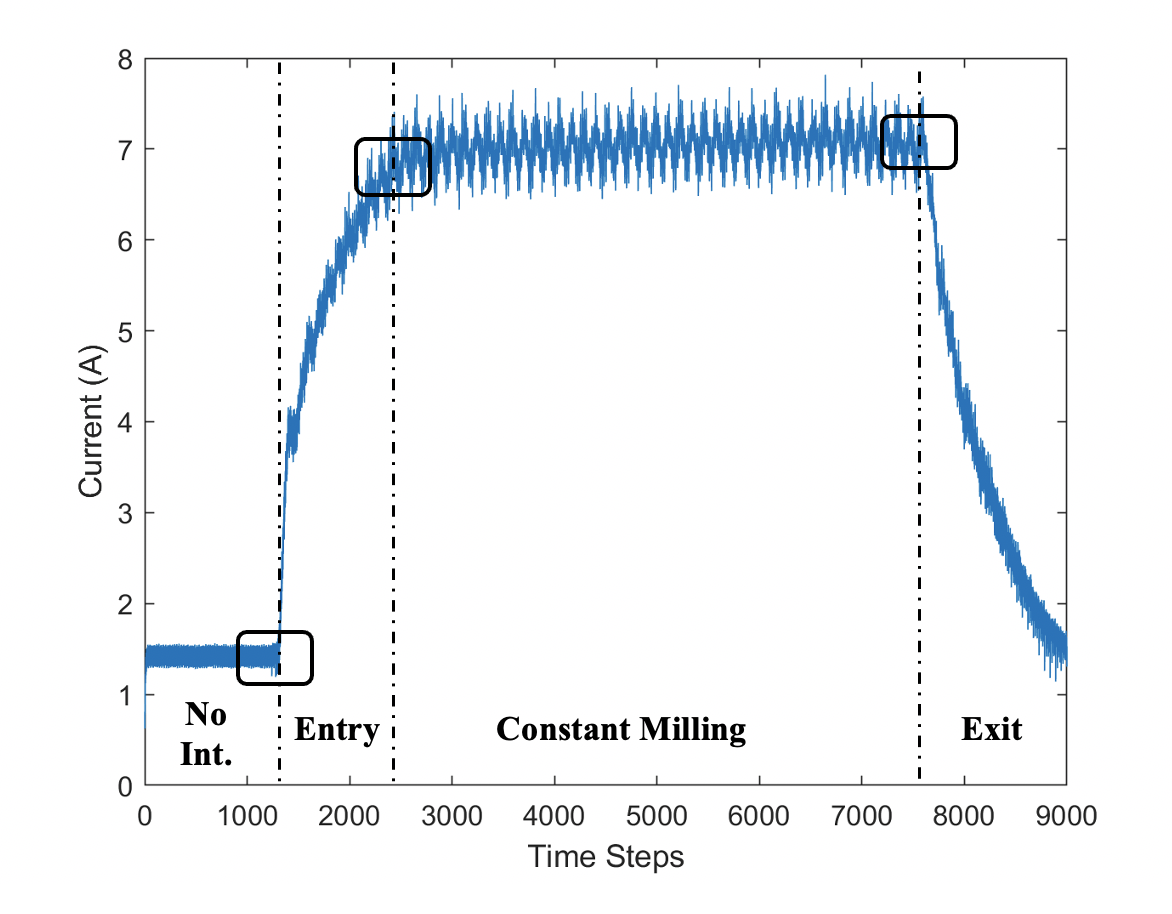}
    \caption{SAMPLE MILLING TRIAL FROM NASA MILLING DATASET}
    \label{fig:dc_sample}
\end{figure}

Based on the progression of interactive states presented in the milling trials, three transitioning events that trigger the following transitions: no interaction to entry transition, entry to constant milling transition, and constant milling to exit transition, are considered. A typical DC spindle current signal, with interactive states labeled and transitions indicated with rectangles, is presented in Fig. \ref{fig:dc_sample}.

\subsection*{Framework Implementation}

The signal processing unit is implemented to partition the incoming stream of the DC current signal into segments of 400 data points with each segment overlapping the preceding and succeeding segment by 25 data points. Eight successive segments are packaged into an input to the classification unit. 

The classification unit accepts inputs of size $8 \times 1 \times 400$. The upstream CNN encodes the input into 8 4-class scores (4 interactive states) of size $8 \times 4$, which are then fed into the downstream CNN. A 7-class score (4 interactive states + 3 transitioning events) of size $1 \times 7$ is produced as the final output of the classification unit.

The upstream CNN is pretrained with 2936 samples, of length 400 data points, extracted from well-defined steady-state regions. Since the upstream CNN is significantly larger than its downstream counterpart, pretraining provides stability and faster convergence in training.

The entire classification unit is trained end-to-end with the pretrained upstream CNN and randomly initialized downstream CNN. The entire dataset used in this step contains 3576 samples of length 3200 data points. Optimizer and loss function used in this case study are Adam \cite{adam} and Cross-Entropy loss with Softmax function \cite{cross-entropy}, respectively. Architectural details of the classification unit are presented in Tab. \ref{tab:model_architecture} in Appendix A. 

\begin{figure}[t]
    \centering
    \includegraphics[width=0.8\linewidth]{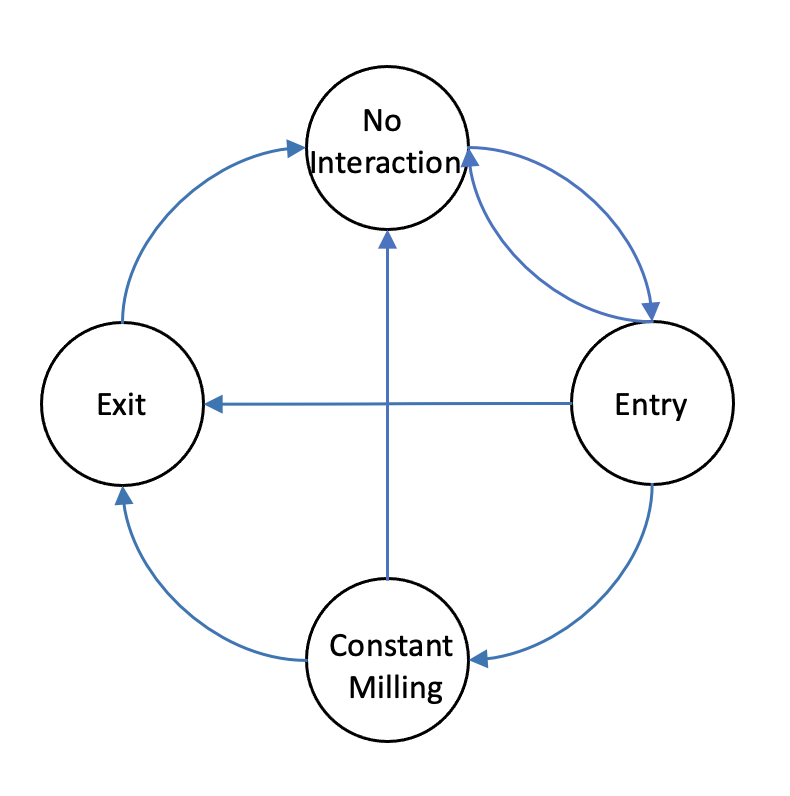}
    \caption{DECISION COORDINATOR STATE TRANSITION DIAGRAM}
    \label{fig:fms}
\end{figure}

The decision coordinator is implemented as a FSM of four states: no interaction, entry, constant milling, and exit. There are seven permissible state transitions, but only three of them exist in the dataset. The state transitioning diagram of the decision coordinator is presented in Fig. \ref{fig:fms}. 

\subsection*{Evaluation}

We use a testing dataset and deployment simulations to evaluate the effectiveness of the implemented system. Performance on the testing dataset quantifies the classification unit's ability to identify interactive state/transitioning events at a given time instance. However, the performance does not provide any insights into the robustness of the system, nor can it quantify delays in indicating the transition to the next interactive state. To address such limitations, we perform deployment simulations, through which we can gauge how the system would perform in real-time.

The testing dataset consists of 894 samples of size $8 \times 1 \times 400$. The samples are constructed in the same manner as the training samples, and they reflect the interactive state of the machine or indicate a transitioning event at a time instance. We use precision, recall, and F1-Score to evaluate the performance of the classification unit. 

For deployment simulations, the DC current signal is streamed as in real-time. The result from each simulation is a series of classifications representing the state of the machine or a transitioning event. The simulation results are compared with the actual state and occurrences of transition to determine the correctness of classifications and delays in detecting transitions. 

\subsection*{Results and Discussion}

We first present the results from the testing dataset, as shown in Tab. \ref{tab:classification_metrics}. The classification unit achieves an F1-Score above 0.9 for all but one class: entry/const transitioning event class. This indicates that the classification unit misclassifies the entry/const transitioning event class more often than other classes. This is expected because entry/const transitioning events are more gradual than other transitioning events, and more data is required to capture their full characteristics. In other words, this class is more difficult to classify than other classes, given a fixed time interval. 

\begin{table}[t]
\caption{PERFORMANCE MEASURES OF THE CLASSIFICATION UNIT}
\centering
\begin{tabular}{c l l l}
\hline
Classes                 & Precision & Recall   & F1-Score  \\ \hline
No Int.              & 0.974  & 1.000        & 0.987 \\
Entry             & 0.992  & 0.959 & 0.975    \\
Const.             & 0.979  & 0.995 & 0.987 \\
Exit              & 0.919  & 0.971 & 0.944 \\
No Int./Entry Tran.  & 0.954  & 0.977 & 0.966 \\
Entry/Const. Tran. & 0.896  & 0.796 & 0.843 \\
Const./Exit Tran.  & 0.956  & 0.896 & 0.925 \\ \hline
\end{tabular}
\label{tab:classification_metrics}
\end{table}

We now present the result of the deployment simulations. The system correctly partitions all five trials into segments that correspond to different interactive states, and intermittent misclassifications only occur in one segment. Since the system is designed for online application, one key performance metric is the delay in identifying the transition to the next interactive state. We manually label the transition points (the time instance when a transition occurs) for the five trials, and the results of the deployment simulations are compared with our labels. Delays, measured in seconds, for each transition are presented in Tab. \ref{tab:delay_table}. 

\begin{table}[t]    
    \centering
    \caption{DELAY IN IDENTIFYING NEW INTERACTIVE STATE IN DEPLOYMENT SIMULATIONS (PROPOSED SYSTEM)}
    \begin{tabular}{c l l l l l}
    \hline
         & Trial 1 & Trial 2 & Trial 3 & Trial 4 & Trial 5  \\ \hline
        
         No Int. to Entry &  0.156 & 0.160 & 0.136 & 0.088 & 0.140 \\
          
          Entry to Const. & -0.700 & 0.820 & 0.380 & 0.036 & 0.324 \\
          
          Const. to Exit & 0.160 & 0.132 & 0.100 & 0.288 & 0.180 \\
          \hline
    \end{tabular}
    \label{tab:delay_table}
\end{table}

It is worthwhile to point out that one of the delays (entry to constant transition of Trial 1) is negative, which means that the detection leads the actual occurrence of the transition. While lagging detections are expected because more context is required beyond the transition point for the classifier to recognize meaningful changes present in the signal, leading detections in general seem unreliable. Such a phenomenon occurs largely due to the fact that entry/const transitioning events are more gradual. Compared to the other two types of transitioning events, entry/const transitioning events are less defined in the sense that a range of data points around the predefined transition point can be viewed as valid transition points.


Finally, we compare our system to standard methods introduced in the background section. To perform such a comparison, we implement a CNN (referred to as the baseline CNN) capable of classifying signals of length 400 as one of the four classes: No interaction, Entry, Constant, and Exit. The baseline CNN can achieve similar performance in the four classes to our system. Since it does not consider transitioning events explicitly, we can only infer occurrences of transition when there is a change in state. The baseline CNN becomes more uncertain as it approaches the transition points, and this is expected because the classifier is only trained for steady-state classes. As a result, it is unable to detect transitioning event consistently, and misclassifications are frequent around the neighborhood of transition points. To quantify the transition detection performance of the baseline CNN, we perform deployment simulations using the baseline CNN. Transition points are manually inferred based on the changes in state, and intermittent misclassifications after the transition points are ignored. The result is presented in Tab. \ref{tab:cnn_delay}.
Our system is able to reliably detect transitioning events and outperforms the baseline CNN in terms of delay in transition detection. It is also able to guard against intermittent misclassifications because of the decision coordinator. Under the restrictions imposed by the decision coordinator, illegal transitions are prevented and the system provides the expected result.

\begin{table}[t]    
    \centering
    \caption{DELAY IN IDENTIFYING NEW INTERACTIVE STATE IN DEPLOYMENT SIMULATIONS (BASELINE CNN)}
    \begin{tabular}{c l l l l l}
    \hline
         & Trial 1 & Trial 2 & Trial 3 & Trial 4 & Trial 5  \\ \hline
        
         No Int. to Entry &  0.356 & 0.36 & 0.448 & 0.396 & 0.448 \\
          
          Entry to Const. & 0.116 & 0.500 & 0.588 & 0.408 & 0.260 \\
          
          Const. to Exit & 0.552 & 0.400 & 0.528 & 0.456 & 0.432 \\
          \hline
    \end{tabular}
    \label{tab:cnn_delay}
\end{table}

\section*{CONCLUSIONS AND FUTURE WORK}
A machine-part interaction classification system is a key component of CPS and can help address a number of manufacturing needs, including anomaly detection, root cause diagnosis, and equipment performance analysis. In this work, a novel framework featuring deep learning techniques combined with Subject Matter Expert knowledge is presented to perform online machine operation classification. The framework is capable of classifying the steady state behaviors of the machine as well as detecting transitions between steady states, and it is validated through a testing dataset and deployment simulations on a milling machine. The deployment simulations illustrate the effectiveness of the SME-designed error checking mechanism. 

In future work, we will investigate applications of the proposed framework to additional case studies. Another potential direction for future work is minimizing the delay in transition detection. Since the system performs classification in real time, minimal delay is always preferred. Currently the framework controls delay through the overlap between successive windows, and one might find it feasible to explore the possibility of using a Recurrent Neural Network-based architecture for the classification unit.

\bibliographystyle{asmems4}
\bibliography{ref}

\appendix       
\section*{Appendix A: Classification Unit Architecture}

\begin{table}[t]
    \centering
    \caption{DETAILED CLASSIFICATION UNIT ARCHITECTURE}
    \scalebox{0.8}{
    \begin{tabular}{c l l l l}
    \hline
    Layer & \multicolumn{2}{l}{Upstream CNN} & \multicolumn{2}{l}{Downstream CNN} \\ \hline
                       & Description     & Parameters & Description     & Parameters   \\ 
1                      & Conv 3$\times$3,5      & 15         & Conv 3$\times$3,8      & 96           \\
2                      & MaxPool,2       &            & BatchNorm      & 8$\times$3          \\
3                      & ReLU            &            & Conv 3$\times$3, 16    & 384          \\
4                      & BatchNorm,5      & 5$\times$3        & BatchNorm,16      & 16$\times$3         \\ 
5                      & Conv 3$\times$3, 25    & 375        & Linear 128$\times$64   & 8192+64      \\
6                      & MaxPool, 2      &            & Linear 64$\times$7     & 448+7        \\
7                      & ReLU            &            &                 &              \\
8                      & BatchNorm,25      & 25$\times$3       &                 &              \\ 
9                      & Conv 3$\times$3,50     & 3750       &                 &              \\
10                     & MaxPool,2       &            &                 &              \\
11                     & ReLU            &        &                 &              \\
12                     & BatchNorm,50      & 50$\times$3          &                 &              \\ 
13                     & Linear 2500$\times$200 & 500000+200 &                 &              \\
14                     & Linear 200$\times$10   & 2000+10    &                 &              \\
15                     & Linear 10$\times$4     & 40+4       &                 &              \\ 
Total                  &                 & 506634     &                 & 9263        \\
\hline
    \end{tabular}}
    \label{tab:model_architecture}
\end{table}

In this section, we provide a brief overview of the network architecture used in the case study. Overall, the network is relatively small and lightweight, compared to the state of the art architectures used in pattern recognition. With only 15 trainable layers and 518597 parameters, the network takes a small amount of time to train end to end (less than 5 minutes with GPU).

The upstream CNN features 3 conventional CONV-POOL-RELU sub-modules with batch normalization layers interweaved and followed by 3 fully connected layers. Slightly different from its upstream counterpart, the downstream CNN omits pooling layers and nonlinear activation layers. We experimented with several hyperparameter settings (number of layers, filter size for the convolutional layers, and the dimension of fully connected layers) and found that the performance of the network is not heavily influenced by the hyperparameters.

Though the detailed architecture is task-specific, we provide a simple heuristic for designing detailed architectures featuring the proposed encoder-classifier architecture. Depending on the number of input signals, the designer can choose to increase or decrease the number of CONV-POOL-RELU sub-modules. Generally, we expect the network to learn more complicated features and feature more convolutional layers, as the number of input signals increases. For the downstream CNN, we recommend using a smaller and shallower architecture, because it only processes the class scores, highly compact and simplified features, received from the upstream CNN.

\end{document}